%% file: main.tex
\documentclass[10pt,twocolumn,letterpaper]{article}

\usepackage{bm}
\usepackage{mathptmx}
\usepackage{amsmath}
\usepackage[NoDate]{currvita}
\usepackage{array}
\usepackage{tabularx}
\usepackage{booktabs}
\usepackage{bm}
\usepackage{ragged2e}
\usepackage{microtype}
\usepackage{graphicx}
\usepackage{caption}  
\usepackage{cuted}    
\usepackage[table, dvipsnames]{xcolor}
\definecolor{darkyellow}{RGB}{204, 153, 0}

\usepackage[pagenumbers]{iccv} 


\definecolor{iccvblue}{rgb}{0.21,0.49,0.74}
\usepackage[pagebackref,breaklinks,colorlinks,allcolors=iccvblue,colorlinks=true,urlcolor=magenta]{hyperref}

\def\paperID{28} 
\def\confName{ICCV}
\def\confYear{2025}

\title{From Coarse to Fine: Learnable Discrete Wavelet Transforms for Efficient 3D Gaussian Splatting}

\author{
Hung Nguyen \quad
An Le \quad
Blark Runfa Li \quad
Truong Nguyen\\
Video Processing Lab, UC San Diego\\
{\tt\small \{hun004, d0le, rul002, tqn001\}@ucsd.edu}
}

\begin{document}
\maketitle
\begin{strip}
    \centering
    \includegraphics[width=\textwidth]{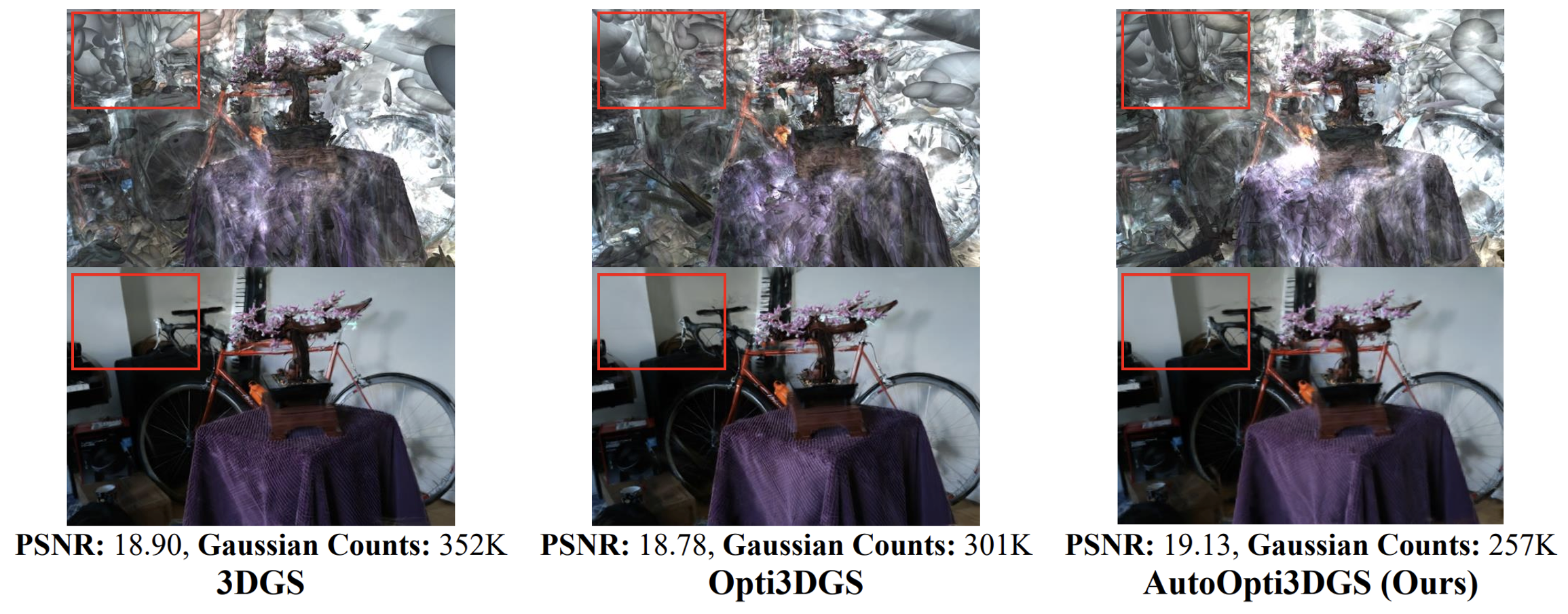}
    \captionof{figure}{We propose AutoOpti3DGS, a framework that optimizes 3DGS \cite{3DGS} Gaussian counts while maintains competitive rendering quality. AutoOpti3DGS works by automatically modulating the input image frequencies in a coarse-to-fine manner, based on the Learnable Discrete Wavelet Transforms. Compared to Vanilla 3DGS \cite{3DGS} and its predecessor Opti3DGS \cite{Opti3DGS}, AutoOpti3DGS reduces the most Gaussian counts, relying on fewer Gaussians to represent relatively homogeneous regions, as exemplified in the red-highlighted region. Top row: Gaussian visualizations. Bottom row: rendered images. The PSNR metric measures pixel-wise differences between rendered and ground-truth images.}
    \label{fig_intro_auto_small}
\end{strip}
\input{sec/0_abstract}    
\input{sec/1_intro}
\input{sec/2_related}
\input{sec/3_prelim}
\input{sec/4_method}
\input{sec/5_exps}
\input{sec/6_conclude}
\section{Acknowledgements}
The first author of this work was financially supported by the Vingroup Science and Technology Scholarship Program for Overseas Study for Master’s and Doctoral Degrees.
{
    \small
    \bibliographystyle{IEEEtran}
    \bibliography{main}
}

\end{document}

%% file: sec/0_abstract.tex
\begin{abstract}

3D Gaussian Splatting has emerged as a powerful approach in novel view synthesis, delivering rapid training and rendering but at the cost of an ever-growing set of Gaussian primitives that strains memory and bandwidth. We introduce AutoOpti3DGS, a training-time framework that automatically restrains Gaussian proliferation without sacrificing visual fidelity. The key idea is to feed the input images to a sequence of learnable Forward and Inverse Discrete Wavelet Transforms, where low-pass filters are kept fixed, high-pass filters are learnable and initialized to zero, and an auxiliary orthogonality loss gradually activates fine frequencies. This wavelet-driven, coarse-to-fine process delays the formation of redundant fine Gaussians, allowing 3DGS to capture global structure first and refine detail only when necessary. Through extensive experiments, AutoOpti3DGS requires just a single filter learning-rate hyper-parameter, integrates seamlessly with existing efficient 3DGS frameworks, and consistently produces sparser scene representations more compatible with memory or storage-constrained hardware. Code is available at: \textcolor{magenta}{\url{https://github.com/hungnguyen311299/AutoOpti3DGS}}.

\end{abstract}

%% file: sec/1_intro.tex
\section{Introduction}
\label{sec:intro}

3D Gaussian Splatting (3DGS) \cite{3DGS} has become a leading approach for reconstructing 3D scenes from collections of 2D images, capable of synthesizing photorealistic views from new perspectives within a reasonable training time. Its strengths have led to widespread applications in fields such as medical imaging \cite{3DGS_Med1, 3DGS_Med2}, robotics \cite{GSSlam, DynaGSSLam}, virtual reality \cite{3DGS_VR1, 3DGS_VR2, SplatSDF, MonoSelfRecon}, and autonomous driving \cite{3DGS_Drive1, 3DGS_Drive2}, among others.

Representing the 3D scene using Gaussians, 3DGS is able to drastically reduce training time and improve rendering quality, compared to its predecessor, NeRF \cite{NeRF}. However, the number of Gaussian primitives is not explicitly controlled during training and tends to grow rapidly as 3DGS adapts to scene details. Therefore, reducing the number of Gaussian primitives while maintaining competitive rendering quality is essential for lowering GPU memory usage and storage demands, especially on edge devices \cite{Compressed-3DGS, Reduced-3DGS}. Fewer Gaussians also benefit methods that rely on per-Gaussian embeddings, which are useful for semantic understanding \cite{LangSplat, LEGaussians}, or for modeling changes in deformation or lighting \cite{WildGaussians, E-D3DGS}.

To this end, we present a novel framework, AutoOpti3DGS. It builds on Opti3DGS \cite{Opti3DGS}, a technique designed to optimize Gaussian counts. Opti3DGS achieves this by applying a coarse-to-fine frequency modulation strategy, progressively sharpening input images during training. Early-stage blurred images lead to larger, coarse Gaussians, which are later splitted into smaller ones that represent fine details as sharper images are introduced. While effective in reducing Gaussian counts without introducing extra parameters or optimization steps, Opti3DGS relies on manually selected blurring parameters, such as filter type (\eg, Gaussian, bilinear, mean), kernel size, and the modulation schedule by which the kernel size is reduced. All of these are not adaptive to the datasets being used, which can compromise rendering quality.

AutoOpti3DGS addresses this limitation by incorporating Learnable Forward and Inverse Discrete Wavelet Transforms (DWT) for adaptive coarse-to-fine modulation of the input images to 3DGS. The Forward DWT decomposes an input image into subbands that capture frequency components at various orientations. The Inverse DWT then reconstructs the image. By fixing low-pass filters and initializing high-pass filters to zero, AutoOpti3DGS starts with coarse reconstructions as input images. An auxiliary loss drives the high-pass filters toward their orthogonal counterparts \cite{wavelet-book}, encouraging high-frequency, fine details to gradually emerge. In summary, the contributions are as follows:
\begin{itemize}
    \item We propose a framework, AutoOpti3DGS, that optimizes Gaussian counts for 3DGS by leveraging Learnable DWT to achieve automatic coarse-to-fine frequency modulation.
    \item Through comprehensive experiments, the proposed AutoOpti3DGS demonstrates its ability to further reduce Gaussian counts while maintaining rendering quality, requiring pre-defining only the learning rate. It also complements well with current efficient 3DGS methods, further reducing Gaussian counts. 
\end{itemize}

%% file: sec/2_related.tex
\section{Related Works}
\label{sec:related}

\textbf{Learnable Discrete Wavelet Transforms}. MLWNet \cite{MLWNet} applies Learnable DWT to discover optimal representations for non-blind motion deblurring. Other works replace pooling layers in CNNs with Learnable DWT to enable adaptive feature extraction. LDW-Pooling \cite{LDW-Pooling} incorporates wavelet-theoretic constraints to prevent filter collapse. uWu \cite{UwU} uses Learnable DWT with orthogonal wavelet initializations for image classification, relaxing the Perfect Reconstruction condition \cite{wavelet-book} for more flexible feature learning. A line of works \cite{Orthogonal-LatticeUwU, Biorthogonal} further extend this framework to orthogonal lattice wavelet structures or biorthogonal wavelets \cite{wavelet-book}, reducing filter complexity or enhancing filter flexibility.

However, while effective for learning CNN-based representations, these methods are incompatible with 3DGS, which does not use CNN features. AutoOpti3DGS bridges this gap by integrating Learnable DWT at the input image level.

\textbf{Discrete Wavelet Transforms for Differentiable Novel View Synthesis.} Recently, wavelets have been utilized in NeRF-based frameworks for different purposes. Rho \etal \cite{MaskedWavelet} uses DWT with learnable masks to sparsify neural radiance fields, reducing memory while preserving rendering quality. WaveNeRF \cite{WaveNeRF} integrates wavelet-transformed multi-scale frequency features into the NeRF pipeline to enhance detail preservation and generalization in novel view synthesis. TriNeRFLet \cite{TriNeRFLet} applies DWT to triplane features for multiscale NeRF representation, enabling coarse-to-fine training, high-frequency sparsity, and super-resolution rendering. DWTNeRF \cite{DWTNeRF} proposes supervising DWT subbands of rendered and ground-truth images for a NeRF variant based on INGP \cite{INGP} encoding to achieve coarse-to-fine detail learning at the loss function level. Regarding 3DGS-based frameworks, MW-GS \cite{MicroMacro} uses the DWT to decompose 2D feature maps into multi-scale frequency components, enabling Gaussians to sample both fine and coarse details. 

However, besides MW-GS, to the best of our knowledge, the use of wavelets in 3DGS is still limited. Our AutoOpti3DGS framework is the first 3DGS method that leverages the DWT to manipulate the input images, with a view to finding an optimal representation that reduces Gaussian counts while maintains rendering quality.

\textbf{Gaussian count optimization.} In Vanilla 3DGS \cite{3DGS}, the Adaptive Density Control (ADC) mechanism reduces or increases the number of Gaussians based on opacity and gradient information. Despite this, Gaussian counts still tend to grow rapidly and unpredictably. To address this issue, Taming-3DGS \cite{Taming-3DGS} introduces a scoring mechanism that combines gradient magnitude, saliency, and internal Gaussian properties to guide densification. This approach enforces a predictable growth curve for the number of Gaussians, enabling models to remain within a specified size budget. Papantonakis et al. \cite{Papantonakis} propose a pruning strategy during training, which successfully reduces Gaussian counts by about 50\%. Mini-Splatting \cite{Mini-Splatting} proposes a Gaussian densification and simplification algorithm where the Gaussians' positions are reorganized, rather than pruning. Compact-3DGS \cite{Compact-3DGS} further reduces redundancy by learning a volume-aware Gaussian mask that eliminates those with minimal rendering contribution. 

However, these approaches all require substantial modifications to the ADC pipeline. Furthermore, their combined effects remain unclear. To complement such methods, Opti3DGS \cite{Opti3DGS} introduces a dataset-level strategy wherein input images are blurred with decreasing intensity throughout training. This encourages the model to form coarse Gaussians early and fine Gaussians later, improving efficiency in Gaussian usage.

However, this approach relies on manually selected blurring parameters that do not adapt to different datasets, which can compromise rendering quality despite reducing Gaussian counts. In contrast, our proposed AutoOpti3DGS replaces hand-crafted blurring based on the Learnable DWT, automatically modulating the input image frequencies in a coarse-to-fine and dataset-adaptive manner.

%% file: sec/3_prelim.tex
\section{Preliminary Background} \label{sec:prelim}

\subsection{3D Gaussian Splatting}

3DGS~\cite{3DGS} utilizes 3D Gaussians to construct an implicit 3D scene, given a set of multi-view 2D input images of the scene. As the 3D scene is modelled, renders from novel perspectives can be obtained. Each Gaussian point is parameterized by a center position $\bm{\mu}$, opacity $\sigma$, covariance matrix $\boldsymbol{\Sigma}$ and color $\mathbf{c}$. To optimize these parameters, the following differentiable loss is used:
\begin{equation} \label{eq_3dgs_loss}
    \mathcal{L}_{3DGS} = (1-\lambda) \mathcal{L}_1(\mathbf{X}^{gt}, \mathbf{X})+\lambda \mathcal{L}_{D-SSIM}(\mathbf{X}^{gt}, \mathbf{X})
\end{equation}
where $\mathbf{X}^{gt}$ and $\mathbf{X}$ denote the ground-truth and rendered images from the same viewpoint, respectively.  $\mathcal{L}_{1}$ is the pixel-wise mean absolute error loss, while $\mathcal{L}_{D-SSIM}$ is a differentiable SSIM loss function \cite{SSIM} representing perceptual similarity. $\lambda$ is a balancing weight between the two.

\subsection{Discrete Wavelet Transforms}

Given a non-square 2D image $\mathbf{X}$, the Forward DWT provides the four subbands, $\mathbf{X}_{LL}$, $\mathbf{X}_{LH}$, $\mathbf{X}_{HL}$, $\mathbf{X}_{HH}$, as follows:
\begin{equation} \label{eq_fdwt}
\begin{split}
\mathbf{X}_{LL} = \mathbf{L}_0\mathbf{X}\mathbf{L}_1, \quad
\mathbf{X}_{LH} = \mathbf{H}_0\mathbf{X}\mathbf{L}_1, \\
\mathbf{X}_{HL} = \mathbf{L}_0\mathbf{X}\mathbf{H}_1, \quad
\mathbf{X}_{HH} = \mathbf{H}_0\mathbf{X}\mathbf{H}_1
\end{split}
\end{equation}
where $\mathbf{L}_{(\cdot)}$ and $\mathbf{H}_{(\cdot)}$ are the low-pass and high-pass analysis matrices that filter the columns or rows of $\mathbf{X}$, respectively. The subscript $\{0, 1\}$ denotes column or row-wise filtering. Generally, the filtering matrices are constructed by shifting rows or columns of the same 1D analysis filter \cite{wavelet-book}. An example is the Haar low-pass filter $\ell = [1/\sqrt{2}, 1/\sqrt{2}]$ and high-pass filter $h = [-1/\sqrt{2}, 1/\sqrt{2}]$.

\begin{figure}[t]
    \includegraphics[width=\linewidth]{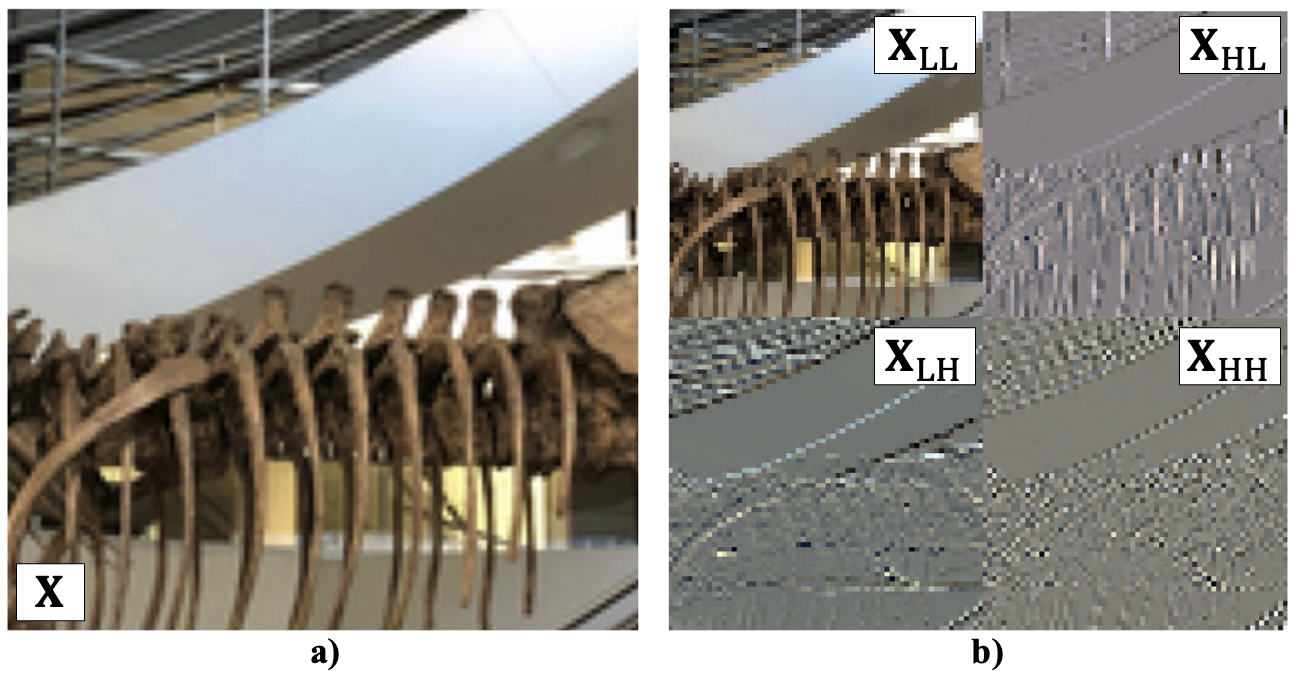}
    \caption{DWT subbands (b) of a cropped region (a) in the ``trex'' scene of the LLFF \cite{LLFF} dataset. The LL subband captures coarse representation. The LH and HL subbands capture horizontal and vertical details, respectively. The HH subband captures diagonal details.}
    \label{fig_vis_subband}
\end{figure}

In Equation \eqref{eq_fdwt}, the filtering matrices are applied across both dimensions. The LL subband results from being low-pass filtered across both dimensions, capturing the image's coarse structure. The LH and HL subbands result from being low-pass filtered along one dimension and high-pass filtered along another, highlighting horizontal or vertical details. Finally, the HH subband results from being high-pass filtered across both directions, capturing diagonal details. A visualization of the DWT subbands is provided in Figure \ref{fig_vis_subband}. 

An additional operation, called the Inverse DWT, reconstructs the original image $\mathbf{X}$ from the subbands:
\begin{equation} \label{eq_idwt}
    \hat{\mathbf{X}} =
    \tilde{\mathbf{L}}_0^\top \mathbf{X}_{LL} \tilde{\mathbf{L}}_1^\top +
    \tilde{\mathbf{H}}_0^\top \mathbf{X}_{LH} \tilde{\mathbf{L}}_1^\top +
    \tilde{\mathbf{L}}_0^\top \mathbf{X}_{HL} \tilde{\mathbf{H}}_1^\top +
    \tilde{\mathbf{H}}_0^\top \mathbf{X}_{HH} \tilde{\mathbf{H}}_1^\top
\end{equation}
where $\mathbf{\hat{X}}$ is the reconstructed image. $\tilde{\mathbf{L}}_{(\cdot)}$ and $\tilde{\mathbf{H}}_{(\cdot)}$ are low-pass and high-pass synthesis matrices, respectively. They are similarly constructed as the analysis matrices, but use 1D synthesis filters $\tilde{\ell}$ or $\tilde{h}$. In the Haar case, the synthesis filters are trivially defined from the analysis filters: $\ell = \tilde{\ell} = [1/\sqrt{2}, 1/\sqrt{2}]$ and $\tilde{h} = [1/\sqrt{2}, -1/\sqrt{2}]$ is the flip of $h$. 

Using a certain set of analysis and synthesis filters, a condition called ``Perfect Reconstruction'' (PR), which occurs when $\mathbf{X} = \mathbf{\hat{X}}$, can be achieved \cite{wavelet-book}. The four Haar wavelet filters used as examples belong to a class called ``orthogonal wavelets'', where orthogonality defines the relationship between low-pass and high-pass filters. Orthogonal wavelets always provide PR.

Our framework focuses on dataset-adaptive, learnable Discrete Wavelet Transforms for automatic coarse-to-fine frequency modulation of input images to 3DGS. We used the implementation of differentiable DWT provided by WaveCNet \cite{WaveCNet}, but additionally derived the gradients w.r.t. the high-pass filters in Section \ref{sec:method}
and in the Supplementary. 



%% file: sec/4_method.tex
\begin{figure*}[t]
    \centering
    \includegraphics[width=0.95\linewidth]{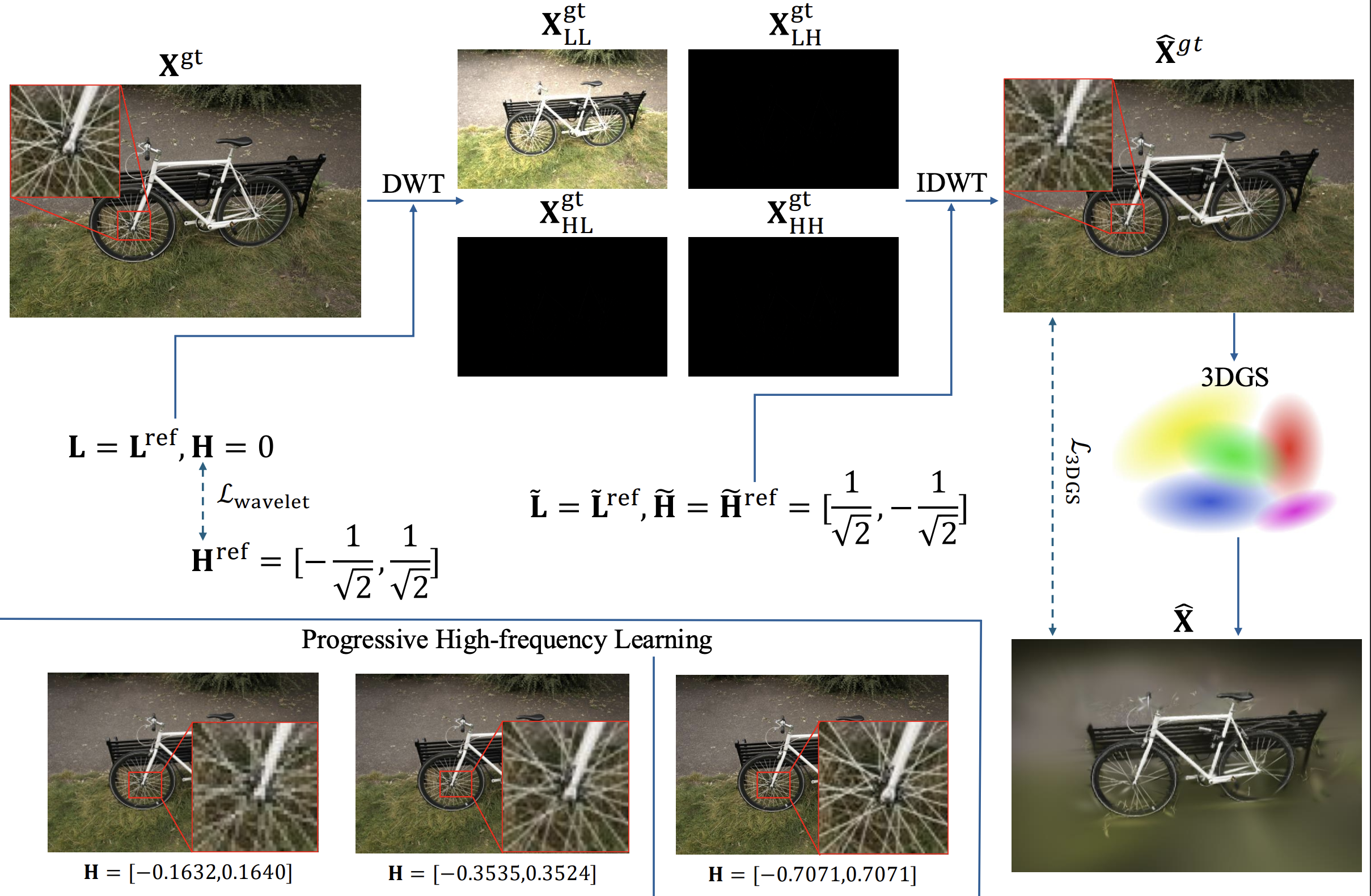}
    \caption{Training overview of AutoOpti3DGS. Coarse-to-fine frequency modulation of the input images is automatically achieved by a succession of Learnable Forward and Inverse Discrete Wavelet Transforms. Under wavelets that satisfy Perfect Reconstruction \cite{wavelet-book}, the process would return perfectly reconstructed images. However, in our framework, the high-pass filter is initialized as zeros, forcing 3DGS to learn from reconstructed coarser images and use only coarse Gaussians. At training time, an auxiliary loss encourages the high-pass filter to converge towards its orthogonal references, introducing finer details and fine Gaussians.}
    \label{fig_autoopti3dgs}
\end{figure*}

\section{Methodology} \label{sec:method}

We present the overview of the AutoOpti3DGS framework in Figure \ref{fig_autoopti3dgs}. Generally, AutoOpti3DGS feeds the input images into the Forward DWT, then the resulting subbands are fed into the Inverse DWT to reconstruct the images. This represents a way of manipulating the input images so that the Gaussian-optimizing representations are achieved.

With regard to coarse-to-fine modulation, the ``coarse'' part is achieved when the input image loses its high frequencies. This can simply be done by initializing the high-pass filter to zeros, which also causes all subbands other than the LL to be a $\mathbf{0}$-matrix. As shown in Figure \ref{fig_autoopti3dgs}, those subbands are, initially, completely black.

To gradually and automatically introduce the high-frequency components, which would accomplish the ``fine'' part of the coarse-to-fine modulation, the high-pass filter must be learnable and adapted to the image. Therefore, the gradients $\partial \mathcal{L} / \partial \mathbf{H}_0$ and $\partial \mathcal{L} / \partial \mathbf{H}_1$ w.r.t. the high-pass analysis matrices in Equation \eqref{eq_fdwt} are additionally derived:
\begin{equation} 
\left\{
\begin{alignedat}{2}
    \frac{\partial \mathcal{L}}{\partial \mathbf{H}_0} 
    &{}= \frac{\partial \mathcal{L}}{\partial \mathbf{X}_{LH}} \frac{\partial \mathbf{X}_{LH}}{\partial \mathbf{H}_0} 
    + \frac{\partial \mathcal{L}}{\partial \mathbf{X}_{HH}} \frac{\partial \mathbf{X}_{HH}}{\partial \mathbf{H}_0} \\
    &{}= \frac{\partial \mathcal{L}}{\partial \mathbf{X}_{LH}} \mathbf{L}_1^\top \mathbf{X}^\top 
    + \frac{\partial \mathcal{L}}{\partial \mathbf{X}_{HH}} \mathbf{H}_1^\top \mathbf{X}^\top \\
    \\
    \frac{\partial \mathcal{L}}{\partial \mathbf{H}_1} 
    &{}= \frac{\partial \mathcal{L}}{\partial \mathbf{X}_{HL}} \frac{\partial \mathbf{X}_{HL}}{\partial \mathbf{H}_1} 
    + \frac{\partial \mathcal{L}}{\partial \mathbf{X}_{HH}} \frac{\partial \mathbf{X}_{HH}}{\partial \mathbf{H}_1} \\
    &{}= \mathbf{X}^T \mathbf{L}_0^T \frac{\partial \mathcal{L}}{\partial \mathbf{X}_{HL}} 
    + \mathbf{X}^T \mathbf{H}_0^T \frac{\partial \mathcal{L}}{\partial \mathbf{X}_{HH}}
\end{alignedat}
\right.
\end{equation}

The high-pass synthesis matrices can be learnt in a similar fashion. In the Supplementary, we conducted an ablation study that showed learning either high-pass analysis or synthesis filters achieves stable frequency modulation. We do not attempt to learn the low-pass filters to conserve the coarse and foundational structure provided by the LL subband. In contrast, learning the high-pass filters poses less risk, as they primarily capture fine details that do not interfere structurally.  

\begin{table*}
\centering
\renewcommand\tabularxcolumn[1]{>{\RaggedRight\arraybackslash}p{#1}}
\begin{tabularx}{\linewidth}{lccccc}
\toprule
&\multicolumn{1}{X}{PSNR ($\uparrow$)}
&\multicolumn{1}{X}{SSIM ($\uparrow$)}
&\multicolumn{1}{X}{LPIPS (VGG) ($\downarrow$)}
&\multicolumn{1}{X}{Peak \#G ($\downarrow$)}
&\multicolumn{1}{X}{Training Time ($\downarrow$) (s)}
\\
\midrule
Opti3DGS \cite{Opti3DGS} & 19.59 & 0.660 & 0.228 & 247K & \textbf{105} \\
AutoOpti3DGS & 20.39 & 0.700 & 0.215 & \textbf{224K} & 133 \\
3DGS \cite{3DGS} & \textbf{20.40} & \textbf{0.706} & \textbf{0.197} & 272K & 109 \\
\midrule
Compact-Opti3DGS \cite{Opti3DGS, Compact-3DGS} & 18.85 & 0.655 & 0.252 & 109K & \textbf{121} \\
Compact-AutoOpti3DGS & 19.83 & 0.681 & 0.229 & \textbf{98K} & 160 \\
Compact-3DGS \cite{Compact-3DGS} & \textbf{19.87} & \textbf{0.682} & \textbf{0.221} & 123K & 139 \\
\midrule 
Mini-Splatting-Opti3DGS \cite{Opti3DGS, Mini-Splatting} & 19.35 & 0.680 & 0.239 & 43K & \textbf{107} \\
Mini-Splatting-AutoOpti3DGS & 20.21 & 0.709 & 0.211 & \textbf{38K} & 134 \\
Mini-Splatting \cite{Mini-Splatting} & \textbf{20.37} & \textbf{0.711} & \textbf{0.203} & 49K & 113 \\
\bottomrule
\end{tabularx}
\vspace{2mm} 
\caption{Comparison of AutoOpti3DGS against Opti3DGS and Vanilla 3DGS, under 3 views of the LLFF \cite{LLFF} dataset}
\label{tab_results_auto3dgs_llff}
\end{table*}

\begin{table*}
\centering
\renewcommand\tabularxcolumn[1]{>{\RaggedRight\arraybackslash}p{#1}}
\begin{tabularx}{\linewidth}{lccccc}
\toprule
&\multicolumn{1}{X}{PSNR ($\uparrow$)}
&\multicolumn{1}{X}{SSIM ($\uparrow$)}
&\multicolumn{1}{X}{LPIPS (VGG) ($\downarrow$)}
&\multicolumn{1}{X}{Peak \#G ($\downarrow$)}
&\multicolumn{1}{X}{Training Time ($\downarrow$) (s)}
\\
\midrule
Opti3DGS \cite{Opti3DGS} & 19.19 & 0.552 & 0.360 & 636K & \textbf{151} \\
AutoOpti3DGS & 19.24 & 0.537 & 0.388 & \textbf{566K} & 177 \\
3DGS \cite{3DGS} & \textbf{19.30} & \textbf{0.564} & \textbf{0.352} & 701K & 155 \\
\midrule 
Compact-Opti3DGS \cite{Opti3DGS, Compact-3DGS} & 18.67 & 0.532 & 0.389 & 262K & \textbf{200} \\
Compact-AutoOpti3DGS & \textbf{18.87} & 0.540 & 0.380 & \textbf{239K} & 222 \\
Compact-3DGS \cite{Compact-3DGS} & 18.85 & \textbf{0.544} & \textbf{0.378} & 301K & 208 \\
\midrule 
Mini-Splatting-Opti3DGS \cite{Opti3DGS, Mini-Splatting} & 18.97 & 0.571 & 0.370 & 107K & \textbf{150} \\
Mini-Splatting-AutoOpti3DGS & 19.20 & \textbf{0.575} & \textbf{0.362} & \textbf{94K} & 188 \\
Mini-Splatting \cite{Mini-Splatting} & \textbf{19.25} & 0.570 & 0.366 & 118K & 160 \\
\bottomrule
\end{tabularx}
\vspace{2mm} 
\caption{Comparison of AutoOpti3DGS against Opti3DGS and Vanilla 3DGS, under 12 views of the Mip-NeRF 360 \cite{MipNeRF360} dataset}
\label{tab_results_auto3dgs_mipnerf}
\end{table*}

To optimize the high-pass filter, the following auxiliary wavelet loss is introduced:
\begin{equation}
\begin{split}
    \mathcal{L}_{wavelet} = \sum_{(\cdot) \in \{0,1\}} ||\mathbf{H_{(\cdot)}} - \mathbf{H}^{ref}_{(\cdot)}||^2_2
\end{split}
\end{equation}
where $(\cdot) \in \{0,1\}$ denotes the vertical or horizontal filtering dimension, and $\mathbf{H_{(\cdot)}}$ denotes the high-pass analysis matrices being learnt. $\mathbf{H}^{ref}_{(\cdot)}$ is the reference with which the images are perfectly reconstructed, \ie, having all fine details. This reference is built upon the Haar 1D high-pass filter introduced in Section \ref{sec:prelim}. Thus, the vertical-filtering, high-pass analysis reference matrix $\mathbf{H}^{ref}_0$ is:
\begin{equation}
\mathbf{H}^{ref}_0 = \begin{bmatrix}
-\frac{1}{\sqrt{2}} & \frac{1}{\sqrt{2}} & 0 & 0 & 0 & \cdots \\
0 & 0 & -\frac{1}{\sqrt{2}} & \frac{1}{\sqrt{2}} & 0 & \cdots \\
0 & 0 & 0 & 0 & -\frac{1}{\sqrt{2}} & \cdots \\
\vdots & \vdots & \vdots & \vdots & \vdots & \ddots
\end{bmatrix} \\
\end{equation}
and similarly for $\mathbf{H}_1^{ref}$ which shifts its 1D filter along columns. As those references are constructed from orthogonal wavelets, convergence of the learnable high-pass filters towards their references implies convergence towards the PR condition, where the high frequencies become available, signalling the end of the coarse-to-fine strategy. Note how the Haar reference filters have only two coefficients, perfectly symmetric in the low-pass case and anti-symmetric in the high-pass case. Therefore, while technically any kind of filter with any number of coefficients can be used, Haar-like filters are easiest to optimize. Please refer to the ``Progressive High-frequency Learning'' section of Figure \ref{fig_autoopti3dgs} to see finer details emerging as training progresses.

\begin{figure*}[ht]
    \centering
    \includegraphics[width=\linewidth]{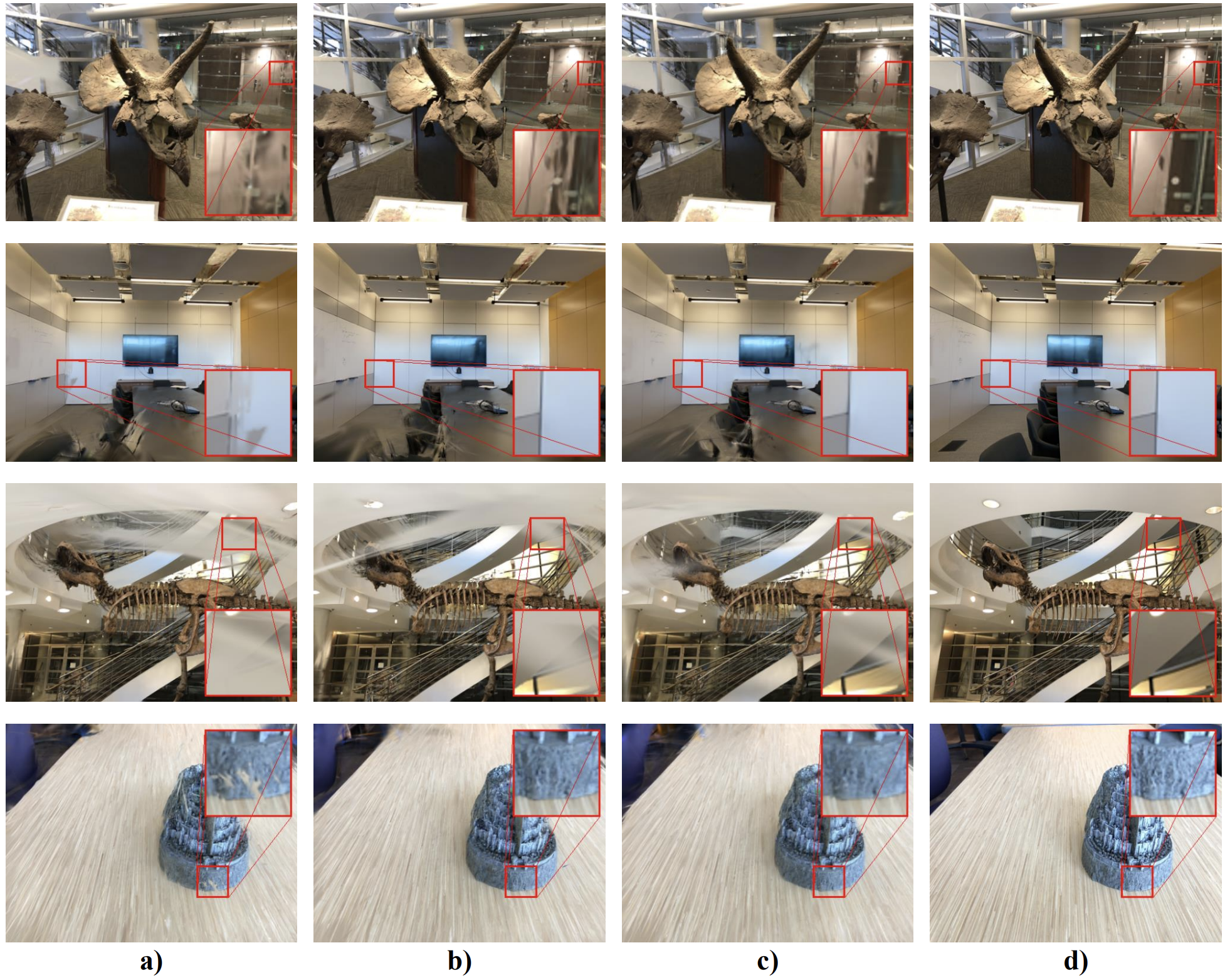}
    \caption{Qualitative results for Opti3DGS (a), 3DGS (b), AutoOpti3DGS (c) and ground-truth (d) on the LLFF \cite{LLFF} dataset (from top to bottom: ``horns'', ``room'', ``trex'' and ``fortress'' scenes). Generally, for this dataset, AutoOpti3DGS is competitive with 3DGS while significantly outperforms Opti3DGS in rendering quality. AutoOpti3DGS also reduces the most Gaussians.}
    \label{fig_quali_autoopti3dgs_llff}
\end{figure*}

The wavelet loss $\mathcal{L}_{wavelet}$ is balanced using the following term:
\begin{equation}
    \lambda_{\mathbf{H}} = \frac{||\mathbf{\hat{X}}_{HH}-\mathbf{X}^{gt}_{HH}||_1}{||\mathbf{X}^{gt}_{HH}||_1 + \epsilon}
\end{equation}
which weighs its importance based on how far the high frequencies of the rendered images ($\mathbf{\hat{X}}_{HH}$) deviate from the high frequencies of the original, fine images ($\mathbf{X}^{gt}_{HH}$). In early training, the HH differences are high, causing the high-pass matrices to converge quickly towards Haar references, and conversely in later training. $\epsilon $ is an extremely small float to prevent division by zero. The total loss for AutoOpti3DGS is as follows:
\begin{equation}
    \mathcal{L} = \mathcal{L}_{3DGS} + \lambda_{\mathbf{H}} \mathcal{L}_{wavelet}
\end{equation}



In summary, AutoOpti3DGS achieves Gaussian count reductions by starting off with coarse images, which result from the high-pass filter being initialized as zeros. Coarse images correspond to coarse Gaussians, because there are little fine details that warrant splitting them. An auxiliary loss encourages the filter to converge towards its references, allowing finer details, or fine Gaussians, to emerge. This optimizes Gaussian counts compared to full-spectrum training from end to end, because in this case, relatively homogeneous regions might be represented by a dense set of fine Gaussians, instead of efficiently represented by a sparse set of coarse Gaussians \cite{Opti3DGS}. Please refer to Figure \ref{fig_intro_auto_small} for a visual example of this.

%% file: sec/5_exps.tex
\begin{figure*}[ht]
    \centering
    \includegraphics[width=\linewidth]{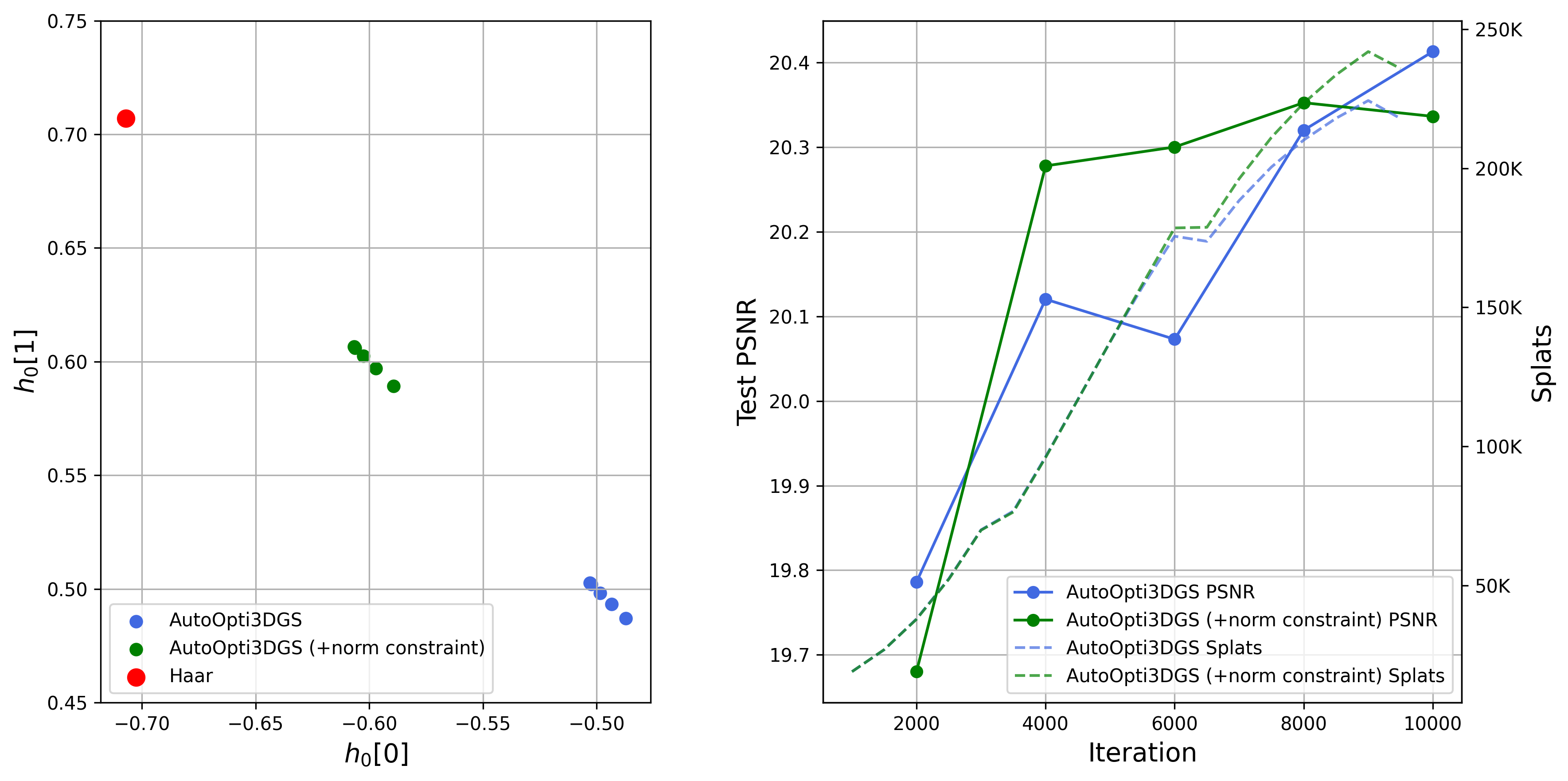}
    \caption{Convergence of vertical high-pass analysis filter (left), as well as PSNR and Gaussian counts at test iterations (right) for two variants of AutoOpti3DGS, evaluated on 3-shot LLFF \cite{LLFF} dataset. The dots  correspond to filter values at iterations 8.5K through 10K, in 0.5K increments. The \textcolor{ForestGreen}{green} AutoOpti3DGS version additionally has a norm constraint when compared to the \textcolor{blue}{blue} version. This pushes the training images closer to PR (left), but incurs more Gaussian counts while possibly decreasing rendering quality (right). This demonstrates the need for frequency selection for 3DGS training images.}
    \label{fig_abs_wavelet_norm}
\end{figure*}

\section{Experiments} \label{sec:exp}

\subsection{Dataset \& Implementation Details}

AutoOpti3DGS is evaluated on the LLFF dataset \cite{LLFF} using 3 input views, and on the Mip-NeRF 360 dataset \cite{MipNeRF360} using 12 input views. These benchmark datasets contain multiple multi-view images of objects or scenes. 3DGS's novel view synthesis capability is tested on some held-out test images that it has never seen during training.

To assess rendering quality, we use PSNR, SSIM \cite{SSIM}, and LPIPS \cite{LPIPS} as evaluation metrics. The first metric measures pixel-wise differences, while the other two measure perceptual similarities. To quantify Gaussian optimization, we track the peak number of Gaussians during training, averaged over all scenes in each dataset. Additionally, we report the training time, also averaged across scenes.

Comparisons are made against Vanilla 3DGS and Opti3DGS, with all three methods incorporating the DropGaussian strategy \cite{dropgaussian}, which randomly drops out Gaussians during training for better generalizability in sparse-view scenarios. Additionally, they are all trained for 10K iterations. For Opti3DGS, blurring is applied using a mean filter with window sizes of 15, 9, 5, and 3 at iteration ranges 0–1K, 1–2K, 2–3K, and 3–4K, respectively. No blurring is used after 4K iterations. These settings follow the original Opti3DGS configuration \cite{Opti3DGS}. In contrast, AutoOpti3DGS only requires setting the learning rate for the high-pass filter, set to 1e-3. 

In addition to Vanilla 3DGS, both Opti3DGS and AutoOpti3DGS are integrated into efficient 3DGS variants to further reduce Gaussian counts. Specifically, we adopt Mini-Splatting \cite{Mini-Splatting} and Compact-3DGS \cite{Compact-3DGS} as base frameworks to showcase the complementary effectiveness of AutoOpti3DGS. All experiments are conducted on a GeForce RTX 4070 Ti Super GPU with 16 GB of memory.

\begin{table*}[t]
\centering
\renewcommand\tabularxcolumn[1]{>{\RaggedRight\arraybackslash}p{#1}}

\begin{tabularx}{\linewidth}{lccccc}
\toprule
&\multicolumn{1}{X}{PSNR ($\uparrow$)}
&\multicolumn{1}{X}{SSIM ($\uparrow$)}
&\multicolumn{1}{X}{LPIPS (VGG) ($\downarrow$)}
&\multicolumn{1}{X}{Peak \#G ($\downarrow$)}
&\multicolumn{1}{X}{Training Time ($\downarrow$) (s)} \\
\midrule
Opti3DGS \cite{Opti3DGS} & 19.59 & 0.660 & 0.228 & 247K & \textbf{105} \\
\rowcolor{yellow}
AutoOpti3DGS - Trainable $\mathbf{H}_{(\cdot)}$ & 20.39 & 0.700 & 0.215 & \textbf{224K} & 137 \\
AutoOpti3DGS - Trainable $\mathbf{H}_{(\cdot)}$ (+$\mathcal{L}_{wavelet\_norm}$) & 20.32 & 0.702 & 0.205 & 242K & 139 \\
\midrule
3DGS \cite{3DGS} & \textbf{20.40} & \textbf{0.706} & \textbf{0.197} & 272K & 109 \\
\bottomrule
\end{tabularx}
\caption{Ablation on wavelet constraints for AutoOpti3DGS, 3-shot LLFF \cite{LLFF} dataset}
\label{tab_abs_constraint}
\end{table*}

\subsection{Quantitative \& Qualitative Results}

Tables \ref{tab_results_auto3dgs_llff} and \ref{tab_results_auto3dgs_mipnerf} summarize the quantitative results. In Table \ref{tab_results_auto3dgs_llff}, across 3DGS and both efficient frameworks, the fixed blurring schedule used in Opti3DGS does not adapt well to the LLFF dataset, resulting in a PSNR drop of $\sim$0.8-1.0, despite reducing Gaussian count by $\sim$10-15\%. In contrast, AutoOpti3DGS maintains rendering quality comparable to 3DGS while reducing the number of Gaussians by $\sim$18-23\%. This improvement comes at some negligible training time, attributed to the use of DWT at every training iteration, but this can likely be mitigated using lazy regularization \cite{StyleGan2}, where the filter updates are performed only every few iterations. Qualitative results on novel views are shown in Figure \ref{fig_quali_autoopti3dgs_llff}.

In Table \ref{tab_results_auto3dgs_mipnerf}, the blurring settings of Opti3DGS perform effectively on the Mip-NeRF 360 dataset, delivering rendering quality on par with 3DGS while reducing Gaussian usage. AutoOpti3DGS achieves similarly competitive performance, while further reducing the number of Gaussians.

\subsection{Ablation study}

This ablation study demonstrates the relative importance of PR convergence for our AutoOpti3DGS framework. The targets for the learnable high-pass filters are orthogonal Haar filters that have unit $\ell_2$-norm, a constraint that has been relaxed due to the learnable nature. To re-introduce the $\ell_2$-norm constraint and make the learnable filter more ``wavelet-like'', the following loss is added into the training objective:
\begin{equation}
    \mathcal{L}_{wavelet\_norm} = \sum_{(\cdot) \in \{0,1\}} \\ \lambda_{\mathbf{H}\_norm} \left(||\mathbf{H_{(\cdot)}}||^2_2 -1\right)
\end{equation}
where $\lambda_{\mathbf{H}\_norm}$ is experimentally set to 0.01 to balance with the other training objectives. The norm loss is only applied after 5K iterations to prevent the high-pass filters from converging too quickly towards the unit-norm references, which would defeat the coarse-to-fine purpose.

The results are displayed in Table \ref{tab_abs_constraint}. Adding $\mathcal{L}_{wavelet\_norm}$ helps the learnable filter converge closer towards its orthogonal references and closer to reconstructing the original images. As shown in the left side of Figure \ref{fig_abs_wavelet_norm}, the final filter using $\mathcal{L}_{wavelet\_norm}$ is $[-0.6062, 0.6061]$, and the final filter not using it is $[-0.5029, 0.5027]$ (the filter values here are the average across rows of the learnable matrix.). The former is closer to the Haar, and thus PR, than the latter. However, this does not result in distinct rendering quality improvements, while requiring larger Gaussian counts, as shown in the right side of Figure \ref{fig_abs_wavelet_norm}. This shows that, at some point, high frequencies stop benefitting. Furthermore, there is a more effective representation than the original input images with regards to Gaussian counts. The Learnable DWT effectively learns this representation. Please refer to the Supplementary for another ablation study on whether to learn the high-pass analysis or synthesis filter. 


%% file: sec/6_conclude.tex
\section{Conclusion}

In this paper, we present the AutoOpti3DGS framework, which aims to reduce the number of Gaussian primitives while preserving rendering quality. Our framework achieves this by enforcing a coarse-to-fine strategy at the input image level, which delays the emergence of fine Gaussians during training. By limiting early-stage supervision to low-frequency content, AutoOpti3DGS prevents the model from prematurely introducing overly fine or redundant Gaussians. This is implemented using Forward and Inverse Discrete Wavelet Transforms (DWT), where the high-pass filter is initialized to zero and gradually activated through training. An auxiliary loss guides it toward its orthogonal reference, enabling progressive reconstruction of fine details. As a result, AutoOpti3DGS eliminates the need for manually defined blurring parameters and only requires setting the wavelet learning rate. Experiments on the LLFF and Mip-NeRF 360 datasets show that AutoOpti3DGS maintains competitive rendering quality while reducing Gaussian counts. Furthermore, it is readily complemented with other efficient 3DGS frameworks, as input image processing does not require major changes. 

\textbf{Future Works:} Firstly, as explained in Section \ref{sec:exp}, we aim to incorporate lazy regularization to the filter updates so as not to incur additional training time, meaning it will take longer for PR convergence. We believe this does not affect the Gaussian reduction benefits of the coarse-to-fine strategy. Secondly, the modulation strategy can be naturally extended to multi-level DWT, where the LL subband from single level is recursively decomposed to produce progressively coarser representations. When the Inverse DWT uses 2-level LL subband as the only non-zero inputs, the resulting reconstructed images will even be coarser than with 1-level LL. We believe this extension has strong potential to further reduce Gaussian counts as it spreads the span of the coarse-to-fine modulation. Additionally, a learning mechanism can be devised for automatically switching DWT levels, from high to low. Nevertheless, without aggressive blurring, AutoOpti3DGS consistently outperforms Opti3DGS in Gaussian count reduction, especially as the latter begins with heavily blurred image inputs mean filtered with large window size (15).